\pdfoutput=1
\documentclass[11pt]{article}

\usepackage[final]{acl}

\usepackage{times}
\usepackage{latexsym}
\usepackage{float}
\usepackage{graphicx}
\usepackage{booktabs}
\usepackage[T1]{fontenc}
\usepackage[utf8]{inputenc}
\usepackage{microtype}
\usepackage{inconsolata}
\usepackage{alphabeta}
\usepackage{ dsfont }
\usepackage{multicol}
\usepackage{amssymb}
\usepackage{amsmath}
\usepackage{multirow}
\usepackage{tikz}

\newcommand*\circled[1]{\tikz[baseline=(char.base)]{
            \node[shape=circle,draw,inner sep=.6pt] (char) {#1};}}
\newcommand{\Dbaby}[1][]{\mathcal{D}_{\text{baby\if\relax\detokenize{#1}\relax\else-\text{#1}\fi}}}
\newcommand{\Dtiny}[1][]{\mathcal{D}_{\text{tiny\if\relax\detokenize{#1}\relax\else-\text{#1}\fi}}}
\newcommand{\Dgen}[1][]{\mathcal{D}_{\text{gen\if\relax\detokenize{#1}\relax\else-\text{#1}\fi}}}

\newcommand{\Ddata}[1][]{\mathcal{D}_{\text{data\if\relax\detokenize{#1}\relax\else-\text{#1}\fi}}}

\newcommand{\Dcomb}[1][]{\mathcal{D}_{\text{comb\if\relax\detokenize{#1}\relax\else-\text{#1}\fi}}}

\newcommand{\strict}{\texttt{Strict}~}
\newcommand{\strictsmall}{\texttt{Strict-Small}~}

\title{BERTtime Stories: Investigating the Role of Synthetic Story Data in Language Pre-training}

\author{Nikitas Theodoropoulos,  {\bf  Giorgos Filandrianos,} {\bf Vassilis Lyberatos},  \\{\bf Maria Lymperaiou} and {\bf Giorgos Stamou}
\vspace{2mm}
\\Artificial Intelligence and Learning Systems Laboratory\\
School of Electrical and Computer Engineering\\ 
\vspace{2mm}
National Technical University of Athens  \\
\texttt{\href{mailto:nikitastheodorop@gmail.com}{nikitastheodorop@gmail.com}},\\
\{\texttt{\href{mailto:geofila@ails.ece.ntua.gr}{geofila}}, \texttt{\href{mailto:vaslyb@ails.ece.ntua.gr}{vaslyb}},
\texttt{\href{mailto:marialymp@ails.ece.ntua.gr}{marialymp}}\}\texttt{@ails.ece.ntua.gr}, \\
\texttt{\href{mailto:gstam@cs.ntua.gr}{gstam@cs.ntua.gr}}
}

\begin{document}
\maketitle

\begin{abstract}
We describe our contribution to the \strict and \strictsmall tracks of the 2nd iteration of the BabyLM Challenge. The shared task is centered around efficient pre-training given data constraints motivated by human development.  In response, we study the effect of synthetic story data in language pre-training using \textsl{TinyStories}: a recently introduced dataset of short stories. Initially,  we train GPT-Neo 
models on subsets of \textsl{TinyStories}, while varying the amount of available data. We find that, even with access to less than 100M words, the models are able to generate high-quality, original completions to a given story, and acquire substantial linguistic knowledge. 
To measure the effect of synthetic story data, we train \textsc{LTG-BERT} encoder models on a combined dataset of: a subset of \textsl{TinyStories}, story completions generated by GPT-Neo, and a subset of the  \textsl{BabyLM} dataset. Our experimentation reveals that synthetic data can occasionally offer modest gains, but overall have a negative influence on linguistic understanding. Our work offers an initial study on synthesizing story data in low resource settings and underscores their potential for augmentation in data-constrained language modeling. We publicly release our models and implementation on our GitHub~\footnote{\href{https://github.com/nikitas-theo/BERTtimeStories}{https://github.com/nikitas-theo/BERTtimeStories}}.
\end{abstract}

\section{Introduction}

As the performance of modern Language Models (LMs) increases, enabling remarkable feats of language understanding and reasoning, so do their demands in computational resources and training data \cite{hoffmann2022trainingcomputeoptimallargelanguage}. For example, the recently released Llama 3 \cite{dubey2024llama3herdmodels} has 405B parameters and was pre-trained on 15.6T tokens, on 6K H100 GPUs. In contrast, 
children are only exposed to no more than 100 million words by age 13 \cite{Jill_Gilkerson_Mapping}, demonstrating exceptional learning efficiency compared to state-of-the-art LMs. This need for ever-increasing data and compute casts doubts on the cognitive plausibility of the current LM training regimes, and raises ecological and ethical concerns, such as democratic access to research for industry and research groups with modest resources. 

To address these issues, the BabyLM challenge \cite{warstadtCallPapersBabyLM2023, choshenCallPapers2nd2024} invites participants to work on cognitive modeling and efficient LM pre-training, given data limitations inspired by human development.  
This year's iteration of the challenge features three experimental tracks: a \strict track with a budget of 100M words, a \strictsmall track with a budget of 10M words, and a 
\texttt{Multimodal} track with a word budget of 100M words and unlimited visual input. A major change compared to last year's challenge 
is allowing participants to construct their own training data. 
In the following sections, we present our contributions to the \strict and \strictsmall tracks.

\begin{figure*}[ht]
\centering
\includegraphics[width=\textwidth]{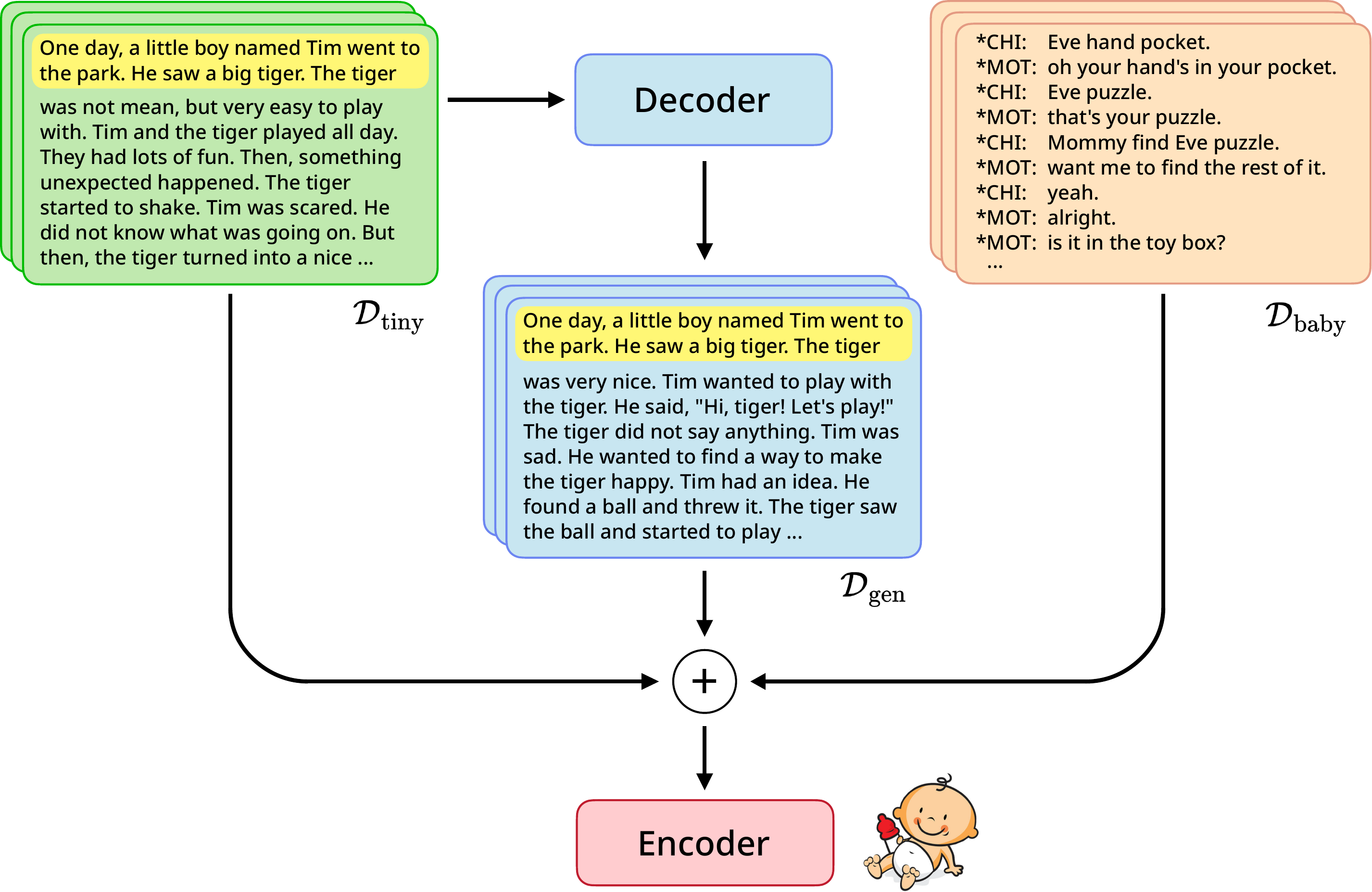}
    \caption{Illustration of our proposed methodology for \textsl{BERTtime Stories}. We use a subset of the \textsl{TinyStories} dataset ($\Dtiny$) \cite{eldanTinyStoriesHowSmall2023}, to train a decoder transformer for data augmentation. We prompt the decoder with the short stories from $\Dtiny$ and create a dataset of model generations $(\Dgen)$: each story (green) is truncated and used as a prompt (yellow), with the model generating an alternate completion (blue). We supplement the two datasets with a subset of the \textsl{BabyLM} dataset ($\Dbaby$), released by \citet{choshenCallPapers2nd2024}, and train an encoder model
    on the combined data. Finally, we evaluate the linguistic proficiency  of the encoder using the challenge benchmarks.
}

\label{fig:pipeline}
\end{figure*}

Our research draws inspiration from recent advancements in Small Language Models (SLMs) for text generation, as explored in \textsl{TinyStories}~\cite{eldanTinyStoriesHowSmall2023}. In this influential work, the authors demonstrate that training on a synthetic dataset of simple stories can enable SLMs to produce creative, high-quality generations, which are novel with respect to the original training dataset.   
We hypothesize that for the small data regimes of the BabyLM challenge, \textit{augmenting} the initial training corpus with synthetic data of high quality can provide models with unseen linguistic contexts, and as a result
improve language understanding.
To test our hypothesis, we first extend previous work 
by \citet{eldanTinyStoriesHowSmall2023}, investigating generative performance with limited training data. We then train encoder transformer models on a diverse dataset, and measure the effect of synthetic data on linguistic proficiency. 


In technical terms, following  \citet{eldanTinyStoriesHowSmall2023}, we propose to train a  
GPT-Neo decoder \cite{Black2021GPTNeoLS}  on \textsl{TinyStories}, and then use it to \textit{generate data} for the training of a final encoder model. This poses some initial challenges, as the size of the original \textsl{TinyStories} dataset exceeds the challenge limits, with around 373M words.  As a result, we experiment with training GPT-Neo models while varying the \textit{amount of available data} and evaluating their generative performance, keeping the model architecture fixed in the process. Our investigation of generative performance is complementary to the original work, which measures the effects of varying the depth and size of the model architectures. Our findings demonstrate that even in low data regimes of $\leq$ 100M words,   GPT-Neo models can acquire impressive grammatical understanding, and generate stories of \textit{high quality} and \textit{diversity}, comparable to models trained on the full dataset. For our evaluation see Section \ref{section:tinystories_gpt_neo}. 

Next, we investigate the effect of the synthetic story data on language pre-training. Figure \ref{fig:pipeline} illustrates our methodology. We select a small subset of \textsl{TinyStories}, train a GPT-Neo model, and use it to generate alternate completions to the stories in the training set. We then create a combined dataset consisting of: \circled{1} the subset of \textsl{TinyStories} used for GPT-Neo training, \circled{2} the generated data, \circled{3} 
a sample of the \textsl{BabyLM} dataset \cite{choshenCallPapers2nd2024}. With the combined dataset, we train an \textsc{LTG-BERT}~\cite{samuelTrained100Million2023b} model, 
choosing this architecture as it demonstrated superior performance in the text-only tracks of last year. We compare the performance of our models against a variety of baselines, trained with and without the use of synthetic data. Our results show that a simple application of synthetic story data for language pre-training results in \textit{marginal} or even \textit{negative} gains. Nevertheless, given the high generative performance of the GPT-Neo models, we believe that more research is needed to fully explore and exploit their potential.

\paragraph{Contribution} We list our contributions below:


\begin{itemize}
    \item We investigate the generative and linguistic abilities of GPT-Neo models trained on \textsl{TinyStories} while varying the amount of available data.
    We show that even with limited data, these models can produce generations of high quality,
    offering new insights into the capabilities of SLMs in low data regimes. 

    \item We investigate the effect of generated data on the pre-training of encoder LMs in a constrained data setting. We conduct an extensive evaluation with different training schemes and baselines. Our experiments demonstrate the potential of data augmentation to enhance the linguistic capabilities of low resource LMs.

\end{itemize}

\section{Related work}

\paragraph{Previous BabyLM Iteration}
Data Augmentation techniques were shown to be beneficial in the previous year's challenge \cite{conll-2023-babylm}. Specifically, ChapGPT \cite{jumelet-etal-2023-chapgtp} uses regex patterns to extract common phrases from GLUE tasks, and then harnesses these patterns to generate follow-up questions that serve as additional training data.  In the Contextualizer paper \cite{xiao-etal-2023-towards}, extra training samples are created by dynamically combining chunks of texts from different contexts during training.  Another approach named Baby's CoThought \cite{zhang-etal-2023-babys} utilizes a Large Language Model (LLM) to reformat unrelated sentences from the corpus into coherent paragraphs, thereby improving performance, albeit in defiance of  data constraints.

\paragraph{Language Models for Data Augmentation}

In recent years, LLMs
have been increasingly leveraged for data augmentation in various domains \cite{ding-etal-2024-data}. Notably, \citet{dai2023auggpt} introduced ChatGPT as a tool for generating realistic text samples from a combination of real and artificial data, enhancing training datasets. Similarly, transformer architectures, including decoder (\textsc{GPT-2}, \citealp{radford2019language}), encoder (\textsc{BERT}, \citealp{devlin-etal-2019-bert}), and seq2seq (\textsc{BART}, \citealp{lewis-etal-2020-bart}) models have been explored for augmentation \cite{kumar2020data}. In the work of \citet{yoo2021gpt3mix},  \textsc{GPT-3} \cite{gpt3} was used to mix real and synthetic text samples for robust data augmentation. Moreover, decoder models 
have been successfully employed to generate training data for encoders,
yielding significant improvements in zero-shot learning 
\cite{mengGeneratingTrainingData2022}. 







\paragraph{Small Language Models}
The recent study by \citet{eldanTinyStoriesHowSmall2023} highlighted that  Small Language Models (SLMs), can outperform larger ones by leveraging high-quality synthetic training data, demonstrating fluency, coherence, and creativity despite having fewer parameters. This trend is further supported by work in sequential recommendation, where small models are effectively employed for task-specific purposes \cite{xu2024slmrec}. Additionally, \citet{bergner2024think} utilize a pre-trained
LLM to encode prompt tokens,
using these representations to guide a smaller LM for more efficient response generation.

\section{Methods}
\label{section:methods}
We describe our data augmentation method  using synthetic story data, as illustrated in Figure~\ref{fig:pipeline}. 
 

\subsection{Datasets}
Our work is built on two datasets: \circled{1} \textsl{TinyStories} -- denoted as $\Dtiny$, a collection of synthetic short stories with simple language, \circled{2} the \textsl{BabyLM} dataset  -- denoted as $\Dbaby$, created to be a developmentally plausible pre-training corpus. 
For any dataset $\Ddata$,  we also denote a version of the data with $m$ million words as $\Ddata[m]$. We describe the datasets below:

\paragraph{\textsl{BabyLM} dataset} 
\phantomsection
\label{paragraph:baby_lm_dataset}
The \textsl{BabyLM} dataset ($\Dbaby$), released by \citet{warstadtCallPapersBabyLM2023, choshenCallPapers2nd2024},
 consists of a diverse set of texts and is constructed with the goal of simulating the linguistic input that a child receives throughout its development. It contains a high proportion of spoken language and includes, among others, excerpts from children's books, dialogue, child-directed speech, and Wikipedia articles. Both 100M and 10M versions of the dataset were released, for the \strict and \strictsmall tracks respectively. Details about the dataset structure are provided 
 in Appendix \ref{appendix:babylm_dataset}. 
\paragraph{\textsl{TinyStories} dataset}  Introduced by \citet{eldanTinyStoriesHowSmall2023},  \textsl{TinyStories} ($\Dtiny$) is a synthetic dataset, featuring a collection of short stories constructed by prompting \textsc{GPT-3.5} and \textsc{GPT-4} \cite{openai2024gpt4technicalreport}. The dataset was created to preserve all the core elements of natural language, such as grammar and reasoning, while exhibiting limited diversity and size. More specifically, the stories are 2-3 paragraphs long and follow simple plots and themes. In addition, the dataset contains a restricted vocabulary and in general is intended to be on the level of understanding of  3-4 year old children. 
The initial version of the dataset (V1), generated by both \textsc{GPT-3.5} and \textsc{GPT-4}, contains approximately 373M words.  A second version (V2) was later released, with stories generated only by GPT-4 and around 440M words. We use this version in all our experiments. 

\subsection{Data Generation}
We describe the creation of the synthetic story dataset $\Dgen$. To generate the data, we first train a decoder model (GPT-Neo) on a subset of \textsl{TinyStories} denoted as $\Dtiny[m]$. We truncate the stories in $\Dtiny[m]$ to construct prompts and generate alternate completions using our model. 

We start by restricting the size $m$ of the subset, taking into account two factors: the need for adequate  \textit{diversity} in the final corpus, and the need to ensure \textit{high-quality} generations. Given the assumption that generation quality scales with dataset size, we want to select a big enough size $m$ for $\Dtiny[m]$ to enable high-quality generations from our trained models. At the same time, we want to leave the necessary room in our word budget for including a sufficiently large portion of the \textsl{BabyLM} dataset in the final training. This will ensure that our models are exposed to both a large vocabulary and a variety of word contexts. Intuitively, we aim to ensure that our pre-training data is diverse, as children learn from multiple sources of input.

To address this trade-off, we sample from \textsl{TinyStories}, creating a collection of subsets of varying sizes, $\Dtiny[m] : { m \in  \{5, 10, 25, 50, 75, 100\}}$M (millions of words). For each subset, we train a GPT-Neo model and evaluate its generative and linguistic abilities. In our evaluation, we leverage metrics for grammatical understanding, diversity, and generation quality; our metrics are introduced in Section \ref{subsec:evaluation}.  For each of the \strict and \strictsmall tracks, we select a subset $\Dtiny[m]$ and a corresponding GPT-Neo model trained on it, based on our evaluation metrics and the above criteria. To construct $\Dgen$, for each story in  $\Dtiny[m]$, we truncate the story to 15\%-30\% of its size and use it to \textit{prompt} the model for generation. We opt for using a smaller proportion of the original story to avoid duplication, given that stories in $\Dtiny[m]$ will already be in the combined corpus for the training of the encoder transformer. 

Regarding the generation process, we experiment with two methods:  greedy decoding and nucleus sampling \cite{holtzman2020curiouscaseneuraltext}. During sampling, we generate $k$ completions from our models for each prompt. To limit repetition between the $k$ generations (and avoid wasting FLOPs), we calculate   Self-BLEU (Section \ref{subsec:evaluation}) for a set of values of $k$, and select the ones that best balance diversity and the total amount of additional training data. 


\subsection{Final Corpus Creation}
For each of the \strict and \strictsmall tracks, we have created $\Dtiny[m]$, and $\Dgen$  as previously described. We now create the combined dataset $\Dcomb$, used to train the encoder transformer. We allocate our remaining word budget to a subset of the \textsl{BabyLM} dataset ($\Dbaby[b]$), created  by sampling randomly from \textsl{BabyLM} on the document level. We leave sampling methods that account for the content of the documents  for future work. For the \strict/~\strictsmall tracks, the size $b$ of $\Dbaby[b]$ is chosen such that: $b + m  \leq 100\text{M}$ / 
$10\text{M}$. We now construct $\Dcomb$ by combining all the datasets  $\Dcomb = (\Dtiny[m],  \Dbaby[b],  \Dgen)$. 
We employ a masked language modeling objective to train an encoder transformer on $\Dcomb$, with the \textsc{LTG-BERT} architecture \cite{samuelTrained100Million2023b}.

\subsection{Evaluation}\label{subsec:evaluation}
For evaluating the encoder transformers we use
the evaluation suite of the challenge, consisting of three evaluation benchmarks:  BLiMP, (Super)GLUE, and EWoK, each broadly evaluating language proficiency, 
general language understanding, and world knowledge. We note that the challenge benchmarks constitute filtered versions \cite{conll-2023-babylm}, rendering our results incomparable with full data evaluations. For the decoder models, we use EWoK and BLiMP, and also introduce some additional evaluation procedures:  specifically, Self-BLEU evaluates diversity, and an LLM-assisted evaluation measures generation quality.
We explain each of the evaluation benchmarks below.

 \paragraph{BLiMP} The Benchmark of Linguistic Minimal Pairs (BLiMP), introduced by \citet{blimp}, is a set of tasks designed to evaluate the linguistic knowledge of LMs. It consists of pairs of minimally different sentences covering various grammatical phenomena in  syntax, morphology, and semantics. The model under evaluation has to assign a higher probability to the 
 correct sentence in each pair.   We also evaluate on BLiMP Supplement (Supp.), released by 
\citet{warstadtCallPapersBabyLM2023}, which includes additional grammatical phenomena. For both BLiMP and BLiMP Supplement,  we measure performance by calculating the average accuracy across all of their evaluation tasks.

\paragraph{(Super)GLUE} The General Language Understanding Evaluation (GLUE) benchmark  \cite{wang2018glue}, assesses model performance across a wide range of natural language understanding (NLU) tasks.  SuperGLUE \cite{NEURIPS2019_4496bf24}, was later introduced to offer a more challenging set of tasks.  We employ a total of 10 text classification tasks from both benchmarks, which include: question answering (BoolQ, MultiRC), sentiment classification (SST-2), paraphrase detection (MRPC, QQP), linguistic acceptability (CoLA), commonsense reasoning (WSC), and  natural language inference (MNLI, QNLI, RTE). Performance on (Super)GLUE is calculated by averaging accuracies across all tasks except for QQP and MRPC, where we use the F1-score, and CoLA, where we use the Matthews Correlation Coefficient -- MCC.


\paragraph{EWoK} Elements of World Knowledge (EWoK) \cite{ivanova2024elements} assesses an LM’s ability to understand and model world knowledge. It evaluates how well a model can connect a target text to either an appropriate or mismatched context, emphasizing key concepts such as social dynamics and spatial relationships. Both the contexts and targets are framed as minimally contrasting pairs, with customizable elements like objects, agents, and locations. During  evaluation, the model needs to assign a higher probability to the correct context and target text pair. We report average accuracy across all the benchmark's tasks.


\paragraph{Self-BLEU} To measure the diversity of generated stories, we utilize the Self-BLEU score ~\cite{zhu2018texygen}. Given a generated collection, we calculate the BLEU score with one generation as the hypothesis and the others as reference, evaluating how similar it is to the rest. We define Self-BLEU as the average of all the BLEU scores in the corpus.  The metric is defined on a continuous scale within $[0,1]$, where higher scores indicate less diversity.

\paragraph{LLM Evaluation} To provide a comprehensive evaluation of our decoder models' generative abilities, we follow the approach of \citet{eldanTinyStoriesHowSmall2023} and employ a LLM, prompting it with the story completions, and asking it to assess them in terms of \emph{Grammar}, \emph{Creativity}, and \emph{Consistency} with the story's beginning, on a scale from 1 to 10. The original evaluation by \citet{eldanTinyStoriesHowSmall2023} used GPT-4,  we instead leverage Claude-3.5 Sonnet~\cite{anthropic_claude}\footnote{Model version: \texttt{claude-3-5-sonnet-20240620}.}, which better aligned with our available resources. 
Evaluation details are presented in Section \ref{paragraph:evaluating_quality_llm},  while the prompt is included in Appendix \ref{sec:LLM_evaluation}.


\section{Experiments}

\paragraph{Experimental Setup} We conduct our experiments on a shared GPU cluster of 8 Nvidia V100 16 GB GPUs, and additionally evaluate our models on an Nvidia RTX-3090 24 GB GPU. All our models are trained using the PyTorch \cite{DBLP:journals/corr/abs-1912-01703} and HuggingFace \cite{DBLP:journals/corr/abs-1910-03771} libraries. For our evaluations of BLiMP, EWoK, and (Super)GLUE we build upon the official evaluation pipeline released by the challenge organizers \cite{eval-harness, choshenCallPapers2nd2024}.

\subsection{TinyStories \& GPT-Neo Evaluation}
\label{section:tinystories_gpt_neo}

Regarding the decoder used for the generation, we select one of the best-performing GPT-Neo architectures from  \citet{eldanTinyStoriesHowSmall2023}~\footnote{\href{https://huggingface.co/roneneldan/TinyStories-33M}{https://huggingface.co/roneneldan/TinyStories-33M}}.  All our trained GPT-Neo models share the same hyperparameters, except for weight decay, dropout, and vocabulary size, which are tuned to the specific data size. We built upon a similar training scheme as the authors, with added regularization for our low data regime. 
Hyperparameters and details about the architecture are included in Appendix \ref{appendix:train-details}. We opt to train on the latest version of the \textsl{TinyStories} data (V2), generated by prompting \textsc{GPT-4}; the full unsampled dataset contains $\sim$ 440M words. Throughout our evaluation, we also report results for the original model released by the authors, trained on the first version of the dataset (V1) with $\sim$ 373M words.

In the following paragraphs, we conduct a thorough analysis of the relationship between the linguistic competency of GPT-Neo models trained on subsets of \textsl{TinyStories}, and  the size of their training dataset $|\Dtiny[m]|$. We experiment with various sizes for the 
\textsl{TinyStories} subsets $\Dtiny[m] : { m \in  \{5, 10, 25, 50, 75, 100\}}$M (millions of words).  
From our experiments we draw insights about the abilities of generative LMs on low data regimes. This evaluation will also motivate our selection of the \textsl{TinyStories} subset $\Dtiny$ used for generating the dataset $\Dgen$ and for training the final encoder.

As an initial proxy of the language competency of the GPT-Neo decoders,  we measure performance on BLiMP, its supplement (Supp.), and EWoK.
Results are presented in Table~\ref{table:gpt_neo_blimp}.  We notice that 50M words appear to be a cutoff point, with notable drops in performance for data sizes less than that. Based on this, we select $\Dtiny[50M]$ for the \strict track, and  $\Dtiny[5M]$ for the \strictsmall track. Importantly, we do not include the LLM evaluation (presented below) in this decision process, as it would invalidate our imposed data constraints. We leave further experimentation on the subset data sizes for the \strictsmall track for future work. A second observation concerns the 100M words model, which achieves the top score on BLiMP, shared by the 373M model by \citet{eldanTinyStoriesHowSmall2023}. This result agrees with the findings of \citet{zhang-etal-2021-need}, demonstrating that 100M words are enough to attain substantial grammatical knowledge.

\begin{table}[h]
    \centering \small
        \begin{tabular}{l|c|c|c}
        \toprule
        Train Data  &  BLiMP $\uparrow$ & Supp. $\uparrow$ & EWoK $\uparrow$\\
        \midrule
        5M & 55.5 & 53.8 & 51.1 \\
        10M  & 58.4 & 51.6 & 51.9\\
        25M  & 59.9 & 55.1 & 52.4\\
        50M & 62.8 & 52.8 & 53.0\\ 
        75M & 64.0  & 54.8 & 53.4\\
        100M  & \textbf{64.8} & 50.8 &  53.1\\
        440M (V2)  & 64.6 & 55.0 & 53.9\\
        373M (V1) \footnotemark & \textbf{64.8} & \textbf{60.9} & \textbf{54.0}\\        
    \bottomrule
    \end{tabular}
    \caption{Evaluation results for GPT-Neo models trained on \textsl{TinyStories} with various amounts of data. We report \textit{accuracy} for all benchmarks. As the amount of data decreases, the BLiMP and EWoK scores generally decrease as well. In contrast, the BLiMP supplement score demonstrates more variance.}
    \label{table:gpt_neo_blimp}
\end{table}

\footnotetext{Model released by \citet{eldanTinyStoriesHowSmall2023}.}

The aforementioned scores give us evidence about the grammatical understanding (BLiMP) and world knowledge (EWoK) of our models, but leave out two important areas of generative performance, mainly: \circled{1} the \textit{diversity} and \circled{2} the \textit{quality} of generations. We focus on these two metrics in the following paragraphs. Apart from the quantitative scores, in Appendix \ref{appendix:tinystories_qualitative}
we also include the generations of all the GPT-Neo models for the \textsl{TinyStories} example illustrated in Figure~\ref{fig:pipeline}.

\paragraph{Evaluating Generation Quality} \label{paragraph:evaluating_quality_llm}
Evaluating the quality of generations for open-ended generation tasks
is challenging, as most common evaluation paradigms expect structured output, and measure fidelity towards a set of reference texts. To address this, we adopt the evaluation method proposed by  \citet{eldanTinyStoriesHowSmall2023}, and prompt an LLM to evaluate the stories generated by our models.  In our experiments, we use Claude-3.5 Sonnet.

We harness a set of 44 manually constructed prompts~\footnote{\href{https://huggingface.co/datasets/roneneldan/TinyStories}{https://huggingface.co/datasets/roneneldan/TinyStories}}  containing the beginning of a story, and generate 10 completions for each of our models, sampling with a temperature of 1. We then provide the LLM with the beginning of the story and the model's completion, and ask it in turn to evaluate the model's response  along three axes: (a) \textit{Grammar}, (b) \textit{Creativity}, and (c) \textit{Consistency} with the beginning of the story. Additionally, 
we ask it to classify the story in different age groups, ranging from 3 (or under) to 16 years old. Scores are given on a scale of 1 to 10, and are averaged across stories and completions. The final results are presented in Table  \ref{table:claude}: we notice that limiting the training data, up to even 25M words, results in a minor decrease of performance across all three metrics. This indicates that the quality of the model generations is retained in the small data regime. Additionally, the 100M words decoder achieves impressive scores in all categories, and outperforms all other models in the \textit{Consistency} metric -- demonstrating that 100M words is enough for robust generative performance.

\paragraph{Evaluating Generation Diversity} To measure diversity, we utilize Self-BLEU \cite{zhu2018texygen}, which has been used before as a measure of the diversity of generated data \cite{holtzman2020curiouscaseneuraltext}. For each model, we sample 100 stories from the training set and truncate them to  15\%-30\%, prompting the model to generate an alternate completion to the story's beginning. When sampling from the model, a greedy decoding strategy is employed.
We report Self-BLEU scores, scaled to $[0,100]$, for the set of 100 completions in Table \ref{table:claude} (higher scores correspond to less diverse generations). Our results indicate that models with limited training data can achieve high diversity, while at the same time maintaining generation quality, as demonstrated by the scores of models trained on 25M and 50M words.

\begin{table}[h]
    \centering \small

        \begin{tabular}{l| c c c | c}
        \toprule
        Train Data  & Gr. $\uparrow$ & Cr. $\uparrow$& 
        Cons. $\uparrow$ & SB $\downarrow$\\
        \midrule
        5M & 4.56& 4.99 & 3.37 & 38.6\\
        10M & 5.31 & 5.34 & 3.98 & 38.3\\
        25M & 6.00 & 5.65 & 4.55 & 34.6\\
        50M & 6.01 & 5.53 & 4.54 & 33.0\\
        75M & 6.08 & 5.50 & 4.49 & 37.1\\
        100M  & 6.17 & 5.57 & \textbf{4.78} &  39.8\\
        440M (V2) & 5.88 & 5.53 & 4.49 & 37.3\\
        373M (V1)   & \textbf{6.24} & \textbf{5.73} & 4.70 & \textbf{29.6}\\
    \bottomrule
    \end{tabular}
    \caption{Results on the evaluation of our models by Claude-3.5 Sonnet. We instruct the LLM to access generative performance along three categories:   Grammar (Gr.),   Creativity (Cr.),  Consistency (Cons.).  We also include Self-BLEU (SB), measuring generation diversity.}
    \label{table:claude}
\end{table}

    \begin{table*}[ht]
        \centering \small

       \begin{tabular}{l|l|l|c|c|c|c|c}
            \toprule
            Model & Training Data & Total &  BLiMP & Supp. & EWoK & GLUE & Avg. \\
            \midrule
            \textsc{LTG-BERT} &  $\Dbaby[10M]$  & 10M & 60.6 & 60.8 & 47.6   & 60.3 & 57.3 \\ 
            \midrule
            BabyLlama & $\Dbaby[10M]$ & 10M & 69.8 & 59.5 & 50.7 & 63.3  & 60.8 \\         
            \midrule

            \multirow{9}{*}{\shortstack{\textsc{LTG-BERT} \\ (\textbf{ours})}}  & $\Dbaby[10M]$ & 10M & 62.8 & \textbf{63.7} & 51.2 & 71.0 & \textbf{62.2}\\
            & $\Dtiny[10M]$ & 10M & 59.8 & 54.2 & \textbf{52.2}  & 67.0 & 58.3\\
            & $\Dtiny[10M] + \Dgen[greedy]$ & 20M & 58.7 & 57.8 & 48.9 & 67.1 & 58.1\\ 

            & $\Dbaby[5M] + \Dtiny[5M]$ & 10M  &62.6 & 60.7 & 51.5 & \textbf{71.2} & 61.5\\ 
            & $\Dbaby[5M] + \Dtiny[5M] + \Dgen[greedy]$ & 15M & 62.1 & 60.2 & 50.4 & 70.6 & 60.8\\
            & $\Dbaby[5M] + \Dtiny[5M] + \Dgen[nucleus-1]$ & 15M & 62.5 & 62.3 & 48.8 & 69.5 & 60.8\\
            & $\Dbaby[5M] + \Dtiny[5M] + \Dgen[nucleus-1] \dagger \star$  & 15M & \textbf{63.2} &  59.3 & 50.4 & 71.1 & 61.0\\
            
            & $\Dbaby[5M] + \Dtiny[5M] + \Dgen[nucleus-5]$ & 33M & 62.4 & 60.1 & 50.7 & 69.4 & 60.6\\
            & $\Dbaby[5M] + \Dtiny[5M] + \Dgen[nucleus-10]$ & 56M & 61.0 & 58.4 & 50.1 & 69.5 & 59.8\\
            \bottomrule            

        \end{tabular}
        \caption{Model performance for the 10M word \strictsmall track.}
        \label{tab:10m-strictsmall}
    \end{table*}

\subsection{Data Generation} 
We now describe the creation of the combined dataset $\Dcomb = (\Dtiny[m], \Dbaby[b], \Dgen)$, leveraged for training an encoder LM. For brevity,  details are given below only for the \strictsmall track; the same process is followed for the \strict track. As discussed in Section~\ref{section:tinystories_gpt_neo}, we choose a subset of 5M words from \textsl{Tinystories} ($\Dtiny[5M]$), and use it to train a GPT-Neo model. This model is then employed to generate the dataset $\Dgen$. We adapt the beginning of each story (15\%-30\%) in the training set $\Dtiny[5M]$ as a prompt, and task the decoder to generate alternative completions.  We experiment with different generation techniques, including greedy generation -- $\Dgen[greedy]$, and nucleus sampling -- $\Dgen[nucleus-k]$, where $k$ is the number of  generations per prompt. Finally, the two datasets are combined with a subset of the \textsl{BabyLM} dataset ($\Dbaby[5M]$), ensuring a total size within the 10M word limit, to form $\Dcomb^{10\text{M}} = (\Dtiny[5M], \Dbaby[5M], \Dgen)$.

\begin{figure}[h]
    \centering    \includegraphics[width=0.45\textwidth]{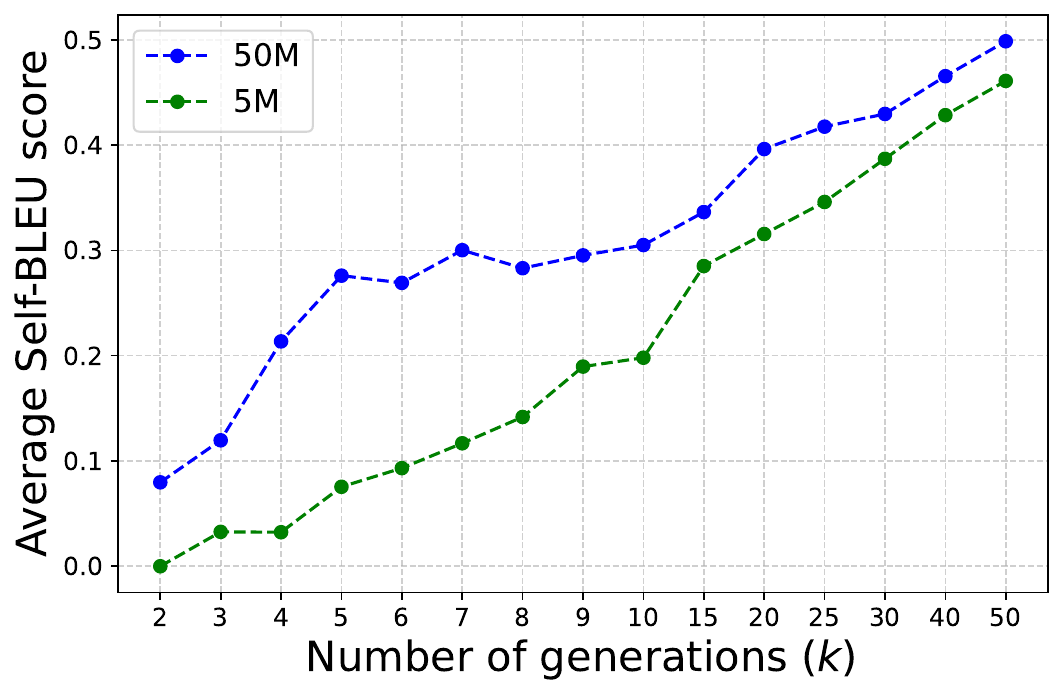}
    \caption{We generate 50 completions for 100 prompts with the GPT-Neo models trained on $\Dtiny[5M]$,  $\Dtiny[50M]$. We plot the average self-BLEU score across prompts, as the number of generations per prompt ($k$) increases.}
    \label{self-bleu}
\end{figure}

In order to select $k$ for nucleus sampling, we leverage the Self-BLEU score. We sample 100 stories from $\Dtiny[5M]$ and use their beginning (15\%-  30\%) to generate 50 completions for each prompt with $p = 0.95$.  For each value of $k \in \{2,3, ..., 50\}$ we calculate Self-BLEU  among the group of generations $S_k$. Our goal is to examine how diverse the different generations are for the same prompt, as the number of generations ($k$) increases. Figure~\ref{self-bleu} depicts the \textit{average} Self-BLEU across all prompts. Based on the presented results, we choose to experiment with $k=5$ and $k=10$, as a satisfactory balance between diversity and added  dataset size.


\subsection{Training \textsc{LTG-BERT}}
\label{section:training_ltg_bert}
Following the creation of the combined corpus $\Dcomb$, we employ it to train an \textsc{LTG-BERT} \cite{samuelTrained100Million2023b} encoder module. Our training procedure is based on the source code released by the authors\footnote{\href{https://github.com/ltgoslo/ltg-bert}{https://github.com/ltgoslo/ltg-bert}}, prompting our selection of similar hyperparameters (Appendix \ref{appendix:train-details}), adapted for our specific infrastructure and available compute. Moreover, our experiments are conducted with minimal hyperparameter optimization. In order to assess the effect of data augmentation on final performance, we train a variety of baselines, ablating over the pre-training dataset of our models and keeping all other training conditions constant. Specifically, for a given track,  all the models share the same hyperparameters and amount of FLOPs, ensuring a fair comparison. Our baselines are described below.

\paragraph{Baselines} For the \strictsmall track, we establish baselines by training \textsc{LTG-BERT} models using 10M words from the \textsl{BabyLM}   -- $\Dbaby[10M]$ and \textsl{Tinystories} -- $\Dtiny[10M]$ datasets respectively. Additionally, we train an encoder using a combination of 5M words from each one of the two datasets -- $\Dbaby[5M] + \Dtiny[5M]$. 
These models serve as benchmarks against which we assess the performance of models trained with various configurations of generated data, aiming to evaluate the effectiveness of data augmentation.  The same methodology is applied consistently to the \strict track as well. Here, we train encoders with 100M words from each dataset separately, as well as in a combined setting, utilizing 50M words from each dataset. 
We also include results for the challenge baselines -- \textsc{LTG-BERT} \cite{samuelTrained100Million2023b} and BabyLlama \cite{timiryasov-tastet-2023-baby}. We emphasize that these models are trained with different hyperparameters than those in our controlled setting. Notably, the \textsc{LTG-BERT} model released by the organizers was trained for $\sim$ 20 epochs on the \strict track, compared to our setting of $\sim$ 27 epochs (20K steps for both tracks).

\begin{table*}[ht]
    \centering \small
    \begin{tabular}{l|l|l|c|c|c|c|c}
        \toprule
        Model & Training Data & Total & BLiMP & Supp. & Ewok & GLUE & Avg\\
        \midrule
        \textsc{LTG-BERT} & $\Dbaby[100M]$ & 100M &69.2 & 66.5 & 50.2 & 68.4 & 63.6\\        
        \midrule
        BabyLlama  & $\Dbaby[100M]$  &  100M & 73.1	& 60.6	&52.1	&69.0	&63.7\\       
        \midrule
        
        \multirow{9}{*}{\shortstack{\textsc{LTG-BERT} \\ (\textbf{ours})}}  & $\Dbaby[100M]$ & 100M & 
        64.0 & \textbf{67.6} & 47.3  & \textbf{74.0} & \textbf{63.2}\\ 
        
        & $\Dtiny[100M]$ & 100M & 61.2 & 63.2 & 48.0 & 70.6 & 60.8\\
        
        & $\Dtiny[100M] +  \Dgen[greedy]$ & 200M & 61.1 & 59.6 & 48.7 & 69.1 & 59.6 \\

        & $\Dtiny[50M] + \Dbaby[50M]$ & 100M & 65.5 & 65.6  & 47.2  &71.0 & 62.3 \\
         
        & $\Dtiny[50M] + \Dbaby[50M]+ \Dgen[greedy]$ & 150M & \textbf{66.6} & 63.3 &  \textbf{49.7} & 71.8 & 62.8 \\
         
        & $\Dtiny[50M] + \Dbaby[50M]+ \Dgen[nucleus-1] \star$ & 150M & 65.6 & 65.0 & 49.3 & 72.7 & 63.1 \\
        
        & $\Dtiny[50M] + \Dbaby[50M]+ \Dgen[nucleus-1] \dagger $ & 150M & 65.2 & 63.5 & 49.0 &  72.6 & 62.6 \\

        & $\Dtiny[50M] + \Dbaby[50M]+ \Dgen[nucleus-5]$ & 350M &65.4 & 64.4 & 45.9 & 69.8 & 61.4\\
         
        & $\Dtiny[50M] + \Dbaby[50M]+ \Dgen[nucleus-10]$ & 600M & 63.7 & 63.3 & 49.2 & 69.5 & 61.4\\

        \bottomrule
    \end{tabular}
    \caption{Model performance for the 100M word \strict track. }
    \label{tab:100m-strict}
\end{table*}


\paragraph{Balanced Training}
While increasing the number of generated texts in the \textsc{LTG-BERT} training set ($\Dcomb$), we also modify the distribution of \textsl{TinyStories} and \textsl{BabyLM} samples that the model encounters during training. This could affect the model's performance, as it becomes more finely tuned to \textsl{TinyStories}. To counter this effect, we experiment with a training variation where we balance the number of samples from both datasets. Specifically, samples in each batch are drawn with equal probability from both \textsl{TinyStories} --  which includes both original  and generated  texts --  and \textsl{BabyLM}. This method ensures that the model is exposed to an equal number of samples from each dataset throughout  training. The dagger symbol $\dagger$ in the results denotes use of this strategy.

\section{Results}

We present the final evaluation results for the  \strictsmall and \strict tracks at Table \ref{tab:10m-strictsmall} and Table \ref{tab:100m-strict}, respectively. The $\star$ symbol denotes the submitted model for this track.

\paragraph{Strict-Small Track}

In the \strictsmall track, comparing the results of 
$\Dbaby[10M]$
with 
$\Dtiny[10M]$
reveals, as expected, that the \textsl{BabyLM} dataset is more beneficial for language pre-training compared to \textsl{TinyStories}.  The performance metrics for \textsl{TinyStories} are consistently lower, except in the case of EWoK. Interestingly, replacing half of the \textsl{BabyLM} dataset with data from \textsl{TinyStories} only slightly affects the model's performance.  However, as we add more instances of the synthetic story data, the positive impact of the \textsl{BabyLM} dataset begins to wane, leading performance to approach that of $\Dtiny[10M]$ where \textsl{BabyLM} was not used at all. This suggests that  training is over-influenced by the increased amount of \textsl{TinyStories} data. To mitigate this effect, we experimented with equally distributing the samples from the two datasets in a batch. This approach positively impacts the model's performance. Notably for BLiMP, this setup 
slightly surpasses the performance of the model trained solely on $\Dbaby[10M]$,  resulting in the best score overall. Further, when compared to  other data augmentation scenarios, the performance on GLUE is increased. 

Moreover, an interesting observation concerns the sampling technique used for augmenting the data. Changing the sampling strategy from greedy decoding to nucleus  sampling positively affects the model's performance on the BLiMP and BLiMP Supp. benchmarks, while negatively impacting performance on EWoK and GLUE. This discrepancy is likely due to the nature of the datasets themselves. BLiMP focuses on evaluating grammatical 
understanding, while the increased diversity from nucleus sampling exposes the model to a wider range of linguistic structures and syntactic variations,
thereby improving performance.
Conversely, EWoK and GLUE require semantic coherence and factual consistency, where the increased diversity from nucleus sampling may introduce noise and less coherent narratives, potentially confusing the model, and degrading performance.
Therefore, while more diverse stories benefit syntactic evaluation tasks such as those in BLiMP, they may not be as useful for semantic or knowledge-based tasks such as those included in EWoK and GLUE.

\paragraph{Strict Track}
Interestingly, for the \strict track we notice that data augmentation has a positive effect on the BLiMP and EWoK benchmarks. Specifically, adding the $\Dgen[greedy]$ dataset, results in increased performance compared to the baselines trained on $\Dtiny[100M]$ and $\Dbaby[100M]$, as well as a mixture of the two ($\Dtiny[50M] + \Dbaby[50M]$).  Additionally, the $\Dtiny[50M] + \Dbaby[50M]$ combination is outperformed by both the $\Dgen[greedy]$ and $\Dgen[nucleus-1]$ models, suggesting that synthetic data can offer modest gains in the \strict scenario.  

As with the  \strictsmall track, increasing the size of the \textsl{TinyStories} dataset negatively affects the performance of the models, approaching that of the model trained solely on $\Dtiny[100M]$. 
However, in this case, balancing the datasets does not improve the model's performance.  In the larger 100M word dataset, even with balancing, the sheer volume of \textsl{TinyStories} data may overwhelm the influence of the \textsl{BabyLM} data. The model is exposed to a much larger quantity of \textsl{TinyStories} content, which could dominate learning and reduce the effectiveness of balancing. Additionally, while the nucleus sampling strategy once again improves performance on the BLiMP Supp. dataset, it does not assist with BLiMP as it did in the \strictsmall track.

\section{Conclusion}
In this work, we explore data augmentation for language pre-training in a 
limited data setting. Using the \textsl{TinyStories} dataset we train GPT-Neo models and probe the relationship between generative ability and dataset size. To measure the effect of augmentation with synthetic data, we train \textsc{LTG-BERT} models on a diverse set of data configurations. Our experiments indicate that while synthesizing high quality data is possible in small data regimes, effectively utilizing it for pre-training can be challenging. Some modest gains are observed in the \strict track, while careful balancing shows promise for the \strictsmall track. 
Overall, our evaluation highlights the intricate balance required between data quantity, quality, and integration for effective training. Future 
work suggests investigation of different data domains, mixtures, and proportions, while precise calibration of hyperparameters may prove critical in exploiting the full benefit of synthetic data  in low data pre-training.



\section{Limitations}
A limitation of our study is the exclusive use of a single LM architecture for both the encoder and decoder components. Our experiments are also limited to specific datasets, employing only \textsl{TinyStories} for synthetic data generation and a combination of \textsl{TinyStories} and \textsl{BabyLM} for encoder training. 
While these choices are made to ensure experimental control and draw solid conclusions, they limit the generalizability of our results.   

Another limitation concerns the creation of the combined dataset. We investigated only a single configuration of the two datasets --  including them in equal proportion -- and the documents within a dataset were sampled randomly. We posit that more fine control over the mixture of datasets could further enhance the benefits of our data augmentation technique. Additionally, with regard to generation, the prompting strategy and truncation ratio could be more finely calibrated, in order to improve the balance between data quality and redundancy.   

By acknowledging these limitations, we aim to encourage further research in this area, focusing on the impact of data augmentation in size constrained and cognitively plausible language pre-training.

\section*{Acknowledgments} The research work was supported by the Hellenic Foundation for Research and Innovation (HFRI) under the 3rd Call for HFRI PhD Fellowships (Fellowship Number 5537).

\bibliography{main}

\onecolumn

\appendix

\section{\textsl{BabyLM} dataset}
\label{appendix:babylm_dataset}

Table \ref{table:babylm_dataset} contains a detailed overview of the  \textsl{BabyLM} dataset. For our experiments, we preprocess the data using the methodology from \citet{samuel-2023-mean}.  The text is normalized and cleaned up in order to ensure a unified format. We cast direct speech in double quotes, remove arbitrary and semantically irrelevant tokens and conserve formatting, where necessary, with a special \texttt{[PAR]} symbol.

\begin{table}[h]
    \centering \small
    \begin{tabular}{llrr}
        \toprule
        \textbf{Dataset} & \textbf{Domain} & \multicolumn{2}{c}{\textbf{\# Words}} \\
        \cline{3-4}
        & & \textbf{Strict-Small} & \textbf{Strict}\\
        \midrule
        CHILDES \cite{MacWhinney2014}  & Child-directed speech & 2.84M & 28.90M\\
        British National Corpus (BNC), dialogue portion $^1$ & Dialogue & 0.93M & 7.76M\\
        Project Gutenberg (children’s stories) \cite{gerlach2018standardizedprojectgutenbergcorpus}  & Written English & 2.54M & 26.37M\\
        OpenSubtitles \cite{lison-tiedemann-2016-opensubtitles2016} & Movie subtitles & 2.04M & 19.96M\\
        Simple Wikipedia $^2$ & Written Simple English & 1.45M & 14.67M\\
        Switchboard Dialog Act Corpus \cite{stolcke-etal-2000-dialogue} & Dialogue & 0.15M & 1.34M\\                
        \midrule
        \textbf{Total} & & \textbf{9.95M} & \textbf{99.01M} \\
        \bottomrule
    \end{tabular}
    \caption{Contents of the \textsl{BabyLM} datasets for the \strict and \strictsmall tracks, including the domain and word counts.  {\small $^1$\href{http://www.natcorp.ox.ac.uk/}{http://www.natcorp.ox.ac.uk/}, $^2$\href{https://dumps.wikimedia.org/simplewiki/20241001/}{https://dumps.wikimedia.org/simplewiki/20241001/}.}}

    \label{table:babylm_dataset}
\end{table}

\section{\textsl{TinyStories} - Detailed Evaluation}
\label{appendix:tinystories_qualitative}
In order to demonstrate a tangible example of the augmentation process, and provide the opportunity to directly judge the quality of the generations, we include sample generations for all our GPT-Neo models: $\{5\text{M}, 10\text{M}, 25\text{M}, 50\text{M}, 75\text{M}, 100\text{M}, 440\text{M} ~\text{(V2)} \}$, as well as the model released by \citet{eldanTinyStoriesHowSmall2023} -- 373M (V1).  We sample a story from the training set, truncate it to around 15\% to 30\% of its length, and ask the models to generate a completion with greedy decoding. The results are shown  in Table~\ref{table:tinygeneration}. The 50M words model generation is also illustrated in Figure~\ref{fig:pipeline}. We notice that even for the smaller models, the quality of the generation remains good in terms of grammar, coherence to the story's beginning, and creativity. Additionally, there is little similarity to the original completion, even though the prompt is taken from the training set of the models.

    \begin{table*}[h]
        \centering \small                \begin{tabular}{p{1.6cm}|p{14cm}}
            \toprule
             \textbf{Model} & \textbf{Generated Text}\\
            \midrule
            Prompt & One day, a little boy named Tim went to the park. He saw a big tiger. The tiger\\
    \midrule
    Original & was not mean, but very easy to play with. Tim and the tiger played all day. They had lots of fun. Then, something unexpected happened. The tiger started to shake. Tim was scared. He did not know what was going on. But then, the tiger turned into a nice dog. Tim was very surprised. Tim and the dog played together now. They were very happy. The dog was easy to play with too. At the end of the day, Tim went home with his new friend.\\
\midrule
    5M & was very big and had a long tail. Tim wanted to be friends with the tiger. Tim
  said, "Hi tiger, I am Tim. Do you want to play with me?" The tiger looked at Tim
  and said, "Yes, I want to play with you." Tim and the tiger played all day. They
  ran, jumped, and laughed. The tiger was very happy. But then, the tiger got tired.
  It wanted to rest. Tim said, "I am tired. Let's rest." The tiger yawned and said,
  "Okay, Tim. Let's rest." Tim and the tiger sat under a tree. They talked and laughed.
  The tiger was happy to have a new friend. They both had a fun day at the park.\\
    \midrule    
    10M &  was very big and had a long tail. Tim was scared, but he wanted to be brave. He walked up to the tiger and said, "Hi, tiger! I am Tim. Do you want to play with me?" The tiger looked at Tim and said, "Yes, I want to play. But I am scared of the big tiger." Tim was not scared. He thought  the tiger was nice. Tim and the tiger played together. They ran and jumped. They had a lot of fun. Then, something unexpected happened. The tiger started to change. It became a tiger! The tiger was not a tiger anymore. It was a tiger! Tim was very surprised. The tiger was not a tiger anymore. It was a big tiger! Tim and the tiger became best friends. They played together every day.\\
    \midrule
    25M & was very big and had a long tail. Tim was scared, but he wanted to be brave. Tim said, "Hi, tiger! I am Tim. Do you want to play with me?" The tiger looked at Tim and said, "Yes, I want to play. But I am very hungry. Can you help me find some food?" Tim thought for a moment and said, "Yes,  I can help you. Let's look for food together." They looked and looked, but they could not find any food. Then, something unexpected happened. The tiger started to talk! "I am not a tiger, I am a magic tiger. I can make your wishes come true," said the tiger. Tim was very surprised. He      wished for a big ice cream cone. The magic tiger made his wish come true. Tim and the magic tiger became good friends and played in the park every day.\\
    \midrule
    50M &  was very nice. Tim wanted to play with the tiger. He said, "Hi, tiger! Let's play!" The tiger
        did not say anything. Tim was sad. He wanted to find a way to make the tiger happy. Tim had an
        idea. He found a ball and threw it. The tiger saw the ball and started to play with it. Tim was
        happy. He said, "Now, the tiger is happy too!" They played together all day. The tiger was not
        sad anymore. Tim and the tiger became best friends.\\
    \midrule
    75M & was very big and had a long tail. Tim was not scared. He wanted to be friends with the tiger. Tim said, "Hi, tiger! My name is Tim. Do you want to play with me?" The tiger looked at Tim and said, "Yes, I want to play. My name is Sam." Tim and Sam played all day. They ran, jumped, and laughed. They had so much fun. Tim was happy to have a new friend. At the end of the day, Tim said, "Bye, Sam! I will come back to play with you again." Sam said, "Bye, Tim! I will wait for you here." Tim went home and told his mom about his new friend.\\
    \midrule
    100M & was sad. Tim asked, "Why are you sad, Mr. Tiger?" The tiger said, "I lost my toy. Can you help me find it?" Tim wanted to help the tiger. They looked for the toy together. They looked under the trees and behind the bushes. Then, they saw a big box. The tiger said, "I think my toy is in the box." Tim opened the box. Inside, there was a big cake! The tiger was happy. But, the cake was not a real tiger. It was a toy tiger! Tim laughed and said, "Silly tiger, you are not a real tiger!" They played with the toy tiger and had a fun day at the park.\\
    \midrule
        440M (V2) & was very hungry. Tim wanted to help the tiger. He thought, "I will feed the tiger some food." Tim went home and got some food. He came back to the park and gave the food to the tiger. The tiger ate the food and was very happy. Tim felt good because he helped the tiger. The tiger said, "Thank you, Tim. You are a good friend." Tim and the tiger played in the park all day. They had lots of fun. Tim learned that it is good to help others. The tiger learned that it is good to be kind and help others. And they both lived happily ever after. \\

\midrule 
373M (V1) & was sad. Tim wanted to help the tiger. He thought the tiger needed a friend. Tim went to the tiger and said, "Hi, I am Tim. I want to help you." The tiger looked at Tim and said, "Thank you, Tim. I am sad because I am alone. I need a friend." Tim thought for a moment and said, "I will be your friend, tiger. We can play together." The tiger was happy. Tim and the tiger played all day. They had lots of fun. The moral of the story is: Helping others can make you happy too. \\

\bottomrule

    \end{tabular}
    
    \caption{
    Story completions for models trained on different amounts of \textsl{TinyStories} data. The original prompt and the training set completion are also included.}
    \label{table:tinygeneration}
\end{table*}

\section{Training Hyperparameters}
\label{appendix:train-details}
We list all the hyperparameters used to train the GPT-Neo models on subsets of \textsl{TinyStories} in Table \ref{table:hyper}. Where  appropriate, parameters that vary per group of models are listed separately. 
We also list the hyperparameters used to train the \textsc{LTG-BERT} transformer encoder models in Table~\ref{table:appendix:ltg-bert-hyperparameters}. We follow the hyperparameter and training regime of \citet{samuelTrained100Million2023b}, with some modifications, partly to adjust to the scale of our computational resources and infrastructure. 

\begin{table}[h]
        \centering
        \small
    \begin{tabular}{l c}
    \toprule
    \textbf{Hyperparameter}  & \textbf{GPT-Neo -- 5 / 10 / 25 / 50, 75, 100, 440 (M)} \\
    \midrule
    Number of Parameters & 41M \\
    Number of Layers &4\\
    Attention Heads  &16\\
    Hidden size & 768\\
    Layer norm $\epsilon$&  1.0e-5\\
    Sequence Length & 512\\
    Max position embeddings & 512\\    
    Attention Dropout & 
    0.50 / 0.40 / 0.25 / 0.20\\
    Classifier Dropout &
    0.50 / 0.40 / 0.25 / 0.20\\    
    Embed. Dropout &
    0.50 / 0.40 / 0.25 / 0.20\\    
    Resid. Dropout &
    0.50 / 0.40 / 0.25 / 0.20\\
    Summary first Dropout &
    0.40 / 0.30 / 0.15 / 0.10\\
    Weight decay & 0.20 / 0.20 / 0.20 / 0.10 \\
    Vocab Size &   6411 / 6411 / 16384 / 16384  \\
    Context length & 512 \\
    batch size & 24\\
    gradient accumulation steps & 32\\
    gradient clipping & 2.0\\
    Training steps & 15 000\\
    optimizer & AdamW\\
    Adam $\beta_1$ & 0.9\\
    Adam $\beta_2$ & 0.95\\
    Adam $\epsilon$ & 1.0e-8\\
    Initial learning rate  & 5.0e-4\\
    Final learning rate  & 5.0e-5\\
    Learning rate scheduler schedule & cosine\\
    Warmup ratio & 1.6\%\\
    
    \bottomrule
    \end{tabular}
    \caption{Hyperparameters used for training GPT-Neo models on \textsl{TinyStories}.}
    \label{table:hyper}
\end{table}

\begin{table}[h]
\centering \small
\begin{tabular}{llll}
\toprule
\textbf{Hyperparameter} & \textbf{Strict} & \textbf{Strict-Small} & \\
\midrule
Number of parameters    &  98M  &  24M  \\ 
Number of layers        & 12            & 12 \\ 
Attention heads         & 12            & 6  \\
Hidden size             & 768           & 384 \\ 
FF intermediate size    & 2048          & 1024  \\
Position Bucket size    & 32            & 32\\
Layer norm $\epsilon$   & 1e-7          & 1e-7\\
Vocabulary size         & 16 384        & 6 144 \\
Sequence length         & 128           & 128   \\
Max position embeddings & 512           & 512\\
Hidden dropout          & 0.1           & 0.1  \\
Attention dropout       & 0.1           & 0.1  \\
Training steps          & 20 000        & 20 000 \\
Batch size              &  80           & 80  \\
Gradient Accumulation Steps & 32        & 32\\
Warmup ratio            & 1.6\%         & 1.6\%  \\
Initial learning rate   & 6e-4          & 6e-4  \\
Final learning rate     & 6e-5          & 6e-5  \\
Learning rate scheduler & cosine        & cosine  \\
Weight decay            & 0.1           & 0.1 \\
Optimizer               & AdamW         & AdamW \\
Adam $\epsilon$         & 1e-6          & 1e-6 \\
Adam $\beta_1$          & 0.9           & 0.9  \\
Adam $\beta_2$          & 0.98          & 0.98 \\
Gradient clipping       & 2.0           & 2.0  \\ 
\bottomrule
\end{tabular}
\caption{Hyperparameters used to train all \textsc{LTG-BERT} models for the different tracks. With \textit{max position embedding} we refer to the architectural capacity of the model -- the model was trained with \textit{sequence length = 128}.}
\label{table:appendix:ltg-bert-hyperparameters}
\end{table}

\section{GLUE Evaluation Hyperparameters}
\label{appendix:glue_eval}
We list the hyperparameters utilized for fine-tuning the \textsc{LTG-BERT} models on GLUE in Table \ref{table:glue_hyperparameters}, largely also based on \citet{samuelTrained100Million2023b}. We find that it's critical for the stability of training to initialize the added classification head according to the original implementation released by the authors. Specifically, we use a truncated normal distribution $\mathcal{N}(\mu, \sigma^2)$ in the range $[a,b]$ where $\mu=0$, $\sigma = \sqrt{\frac{2.0}{5.0 \cdot d_{model}}}$, $a = -2\sigma$, $b = 2\sigma$ and $d_{model}$ is the hidden size. The bias of the layer is also initialized to zero. Lastly, one notable difference affecting the downstream performance of our models is that we train with \textit{mixed precision}. We use the fp16 implementation provided by the HuggingFace Trainer module, which reduces the numerical precision of certain variables from fp32 to fp16, increasing computational efficiency.


\begin{table}[h]
    \centering \small
    \begin{tabular}{l c c}
    \toprule
    \textbf{Hyperparameter} & \textbf{BoolQ, MNLI, MNLI-mm, MRPC} & \textbf{CoLA, RTE, WSC}\\
     & \textbf{ MultiRC, QNLI, QQP, SST-2}\\
    \midrule
    Batch size & 32 & 16\\
    Number of epochs & 10 & 10\\
    Dropot & 0.1 & 0.1 \\
    Warmup proportion & 0.1& 0.1\\
    Learning Rate & 3.0e-5 & 3.0e-5\\
    Decay type & linear & linear\\
    Weight decay & 0.01  & 0.01 \\
    Optimizer & AdamW & AdamW\\
    Adam $\epsilon$ & 1.0e-6 & 1.0e-6\\
    Adam $\beta_1$ & 0.9 & 0.9 \\
    Adam $\beta_2$ & 0.999 & 0.999\\
    Max length & 128 & 128\\

    Gradient clip & 2.0 & 2.0\\
    Mixed precision & True & True\\
    \bottomrule
    \end{tabular}
    \caption{Hyperparameters used for finetuning the \textsc{LTG-BERT} models on GLUE.}
    \label{table:glue_hyperparameters}
\end{table}

\section{LLM Evaluation}
\label{sec:LLM_evaluation}
For the LLM evaluation, we employ Claude-3.5 Sonnet. Our prompt is a modified version of the one introduced in \citet{eldanTinyStoriesHowSmall2023}. We present our prompt in Table \ref{claude:prompt}.

\begin{table}[h] 
    \centering \small
    \begin{tabular}{p{\textwidth}}
\toprule
~\\
In the following exercise, the student is given a beginning of a story. The student needs to complete it into a full story.  The exercise tests the student's language abilities and creativity.  The symbol *** marks the separator between the prescribed beginning and the student's completion:
\\
\\
Once upon a time, there was tiger who liked to play the guitar. One day, a bunny heard the guitar from a distance and ***  asked, "Hello tiger, can you play the guitar for me?" The tiger said, "Yes! I can!" So the bunny plucked up the guitar and started to play. But then, the tiger roared and kicked the guitar too hard. The guitar made a big mess. The panda was so sad that he hopped away and never came back. The end.\\\\Please provide your general assessment about the part written by the student (the one after the *** symbol). Is it gramatically correct? Is it consistent with the beginning of the story? Pay special attention to whether the student manages to complete the sentence which is split in the middle by the separator ***.\\
    \\
{
\itshape
I'll provide an assessment of the student's completion of the story:

Grammar:
The student's portion is generally grammatically correct, with a few minor issues:

...

...

In summary, while the student completes the split sentence correctly, the overall story lacks coherence and consistency with the given beginning. 
}
\\
\\
Now, grade the student's completion in terms of 1. Grammar, 2. Creativity, 3. Consistency with the story's beginning and whether the plot makes sense. Please provide grades from a scale of 1-10 for each of the requested categories, namely: 1. Grammar, 2. Creativity, 3. Consistency. Moreover, please provide your best guess of what the age of the student might be, as reflected from the completion. Choose from possible age groups: A: 3 or under. B: 4-5. C: 6-7. D: 8-9. E: 10-12. F: 13-16.

\\

 \itshape
 Here are the grades for the student's completion on a scale of 1-10:

~\\

\itshape  1. Grammar: 7/10  ...  2. Creativity: 5/10 ...  Consistency: 3/10 ...

~\\

\itshape Age group estimate: B: 4-5 years old   ...

\\
\bottomrule 
\end{tabular}
    \caption{A sample interaction with Claude-3.5 Sonnet. The story is generated by the GPT-Neo model trained on 50M words of \textsl{TinyStories}. We include the prompt which also contains the model's generation. The LLM's responses is stylized  in \textit{italic}.  For brevity,  part of the LLM response was omitted and replaced with `...'.}
    \label{claude:prompt}
\end{table}

\end{document}